\documentclass[11pt,a4paper]{article}
\usepackage[hyperref]{emnlp2018}
\usepackage{times}
\usepackage{latexsym}
\usepackage{url}
\usepackage{graphicx}
\usepackage{amsmath}
\usepackage{amssymb}
\usepackage{subfigure}
\usepackage{caption}
\usepackage{algorithm}
\usepackage{algorithmicx}
\usepackage{algpseudocode}
\usepackage{multicol}
\usepackage{multirow}
\usepackage{booktabs}
\usepackage{listings}

\aclfinalcopy 

\newcommand{\dataset}{FewRel}

\usepackage{siunitx}
\usepackage{todonotes}

\title{{\dataset}: A Large-Scale Supervised Few-Shot Relation Classification Dataset with State-of-the-Art Evaluation}

\author{Xu Han$^{1,}$\thanks{The first four authors contribute equally. The order is determined by dice rolling.}\qquad Hao Zhu$^{1,*}$\qquad Pengfei Yu$^{2,*}$\qquad Ziyun Wang$^{1,}\thanks{Z. Wang is now at New York University.}$$^{\dag,*}$\\ \textbf{Yuan Yao}$^1$\qquad \textbf{Zhiyuan Liu}$^{1,}$\thanks{Correspondence author.}\qquad \textbf{Maosong Sun}$^1$\\ 
\url{http://zhuhao.me/fewrel}
\\
Institute for Artificial Intelligence \\ State Key Laboratory of Intelligent Technology and Systems \\
$^1$Department of CST, $^2$Department of EE, Tsinghua University, Beijing, China
\\ {\tt\{hanxu17,zhuhao15,yupf15,yy18\}@mails.tsinghua.edu.cn,}\\ {\tt ziyunw@nyu.edu, \{lzy,sms\}@tsinghua.edu.cn} 
}

\usepackage{scrextend}
\deffootnote[1.5em]{1.5em}{1em}{\thefootnotemark\space}

\date{}

\begin{document}
\maketitle
\begin{abstract}
We present a Few-Shot Relation Classification Dataset ({\dataset}), consisting of $70, 000$ sentences on $100$ relations derived from Wikipedia and annotated by crowdworkers. The relation of each sentence is first recognized by distant supervision methods, and then filtered by crowdworkers. We adapt the most recent state-of-the-art few-shot learning methods for relation classification and conduct thorough evaluation of these methods. Empirical results show that even the most competitive few-shot learning models struggle on this task, especially as compared with humans. We also show that a range of different reasoning skills are needed to solve our task. These results indicate that few-shot relation classification remains an open problem and still requires further research. Our detailed analysis points multiple directions for future research. All details and resources about the dataset and baselines are released on \url{http://zhuhao.me/fewrel}.


%
%
\end{abstract}

\section{Introduction}

Relation classification (RC) is an important task in NLP, aiming to determine the correct relation between two entities in a given sentence. Many works have been proposed for this task, including kernel methods \cite{Zelenko2002KernelMF,mooney2006subsequence}, embedding methods \cite{Gormley2015ImprovedRE}, and neural methods \cite{Zeng2014RelationCV}. The performance of these conventional models heavily depends on time-consuming and labor-intensive annotated data, which make themselves hard to generalize well. Adopting distant supervision is a primary approach to alleviate this problem for RC \cite{mintz2009distant,riedel2010modeling,hoffmann2011knowledge,surdeanu2012multi,zeng2015distant,lin2016neural}, which heuristically aligns knowledge bases (KBs) and text to automatically annotate adequate amounts of training instances. We evaluate the model proposed by \citet{lin2016neural}, which is followed by the recent state-of-the-art methods \cite{zeng2017incorporating,ji2017distant,huang2017deep,wu2017adversarial,liu2017soft,feng2018reinforcement,zeng2018large}, on the benchmark dataset NYT-10 \cite{riedel2010modeling}. Though it achieves promising results on common relations, the performance of a relation drops dramatically when its number of training instances decrease. About $58\%$ of the relations in NYT-10 are long-tail with fewer than $100$ instances. Furthermore, distant supervision suffers from the wrong labeling problem, which makes it harder to classify long-tail relations. Hence, it is necessary to study training RC models with insufficient training instances.


\text

\begin{table}[t]
\centering
\small
\scalebox{0.89}{
\begin{tabular}{l|p{0.7\columnwidth}}
\toprule
\multicolumn{2}{c}{Supporting Set}\\
\midrule
\multirow{2}{*}{(A) capital\_of}     & (1)  \emph{\textcolor{blue}{London}} is the capital of  \emph{\textcolor{red}{the U.K}}. \\
                                 & (2)  \emph{\textcolor{blue}{Washington}} is the capital of  \emph{\textcolor{red}{the U.S.A}}. \\
\midrule
\multirow{3}{*}{(B) member\_of}      & (1)  \emph{\textcolor{blue}{Newton}} served as the president of  \emph{\textcolor{red}{the Royal Society}}. \\
                                 & (2)  \emph{\textcolor{blue}{Leibniz}} was a member of  \emph{\textcolor{red}{the Prussian Academy of Sciences}}. \\
                                 \midrule
\multirow{4}{*}{(C) birth\_name} & (1) \emph{\textcolor{red}{Samuel Langhorne Clemens}}, better known by his pen name  \emph{\textcolor{blue}{Mark Twain}}, was an American writer. \\
                                 & (2)  \emph{\textcolor{blue}{Alexei Maximovich Peshkov}}, primarily known as  \emph{\textcolor{red}{Maxim Gorky}}, was a Russian and Soviet writer. \\
\midrule
\midrule
\multicolumn{2}{c}{Test Instance}\\
\midrule
\multirow{2}{*}{(A) or (B) or (C)}      &  \emph{\textcolor{blue}{Euler}} was elected a foreign member of \emph{\textcolor{red}{the Royal Swedish Academy of Sciences}}.\\
\bottomrule
\end{tabular}
}
\caption{An example for a 3 way 2 shot scenario. Different colors indicate different entities, \textcolor{blue}{blue} for head entity, and \textcolor{red}{red} for tail entity.}
\label{tab:example}
\end{table}


We formulate RC as a few-shot learning task in this paper, which requires models capable of handling classification task with a handful of training instances, as shown in Table \ref{tab:example}. Many efforts have devoted to few-shot learning. The early works \cite{caruana1995learning,bengio2012deep,donahue2014decaf} apply transfer learning methods to finetune pre-trained models from the common classes containing adequate instances to the uncommon classes with only few instances. Then metric learning methods \cite{koch2015siamese,Vinyals2016MatchingNF-MiniImageNet,Snell2017PrototypicalNF} have been proposed to learn the distance distributions among classes. Similar classes are adjacent in the distance space. The metric methods also take advantage of non-parametric estimation to make models efficient and general. Recently, the idea of meta-learning is proposed, which encourages the models to learn fast-learning abilities from previous experience and rapidly generalize to new concepts. Many meta-learning models \cite{ravi2016optimization,santoro2016meta,finn2017model,Munkhdalai2017MetaN} achieve the state-of-the-art results on several few-shot benchmarks.

Though meta-learning methods develop fast, most of these works evaluate on two popular datasets, Omniglot \cite{Lake2015HumanlevelCL-Omniglot} and mini-ImageNet \cite{Vinyals2016MatchingNF-MiniImageNet}. Both the datasets concentrate on image classification. Many works in NLP mainly focus on the zero-shot/semi-supervised scenario \cite{xie2016representation,Ma2016LabelEF,carlson2009coupling}, which incorporate extra information to classify objects never appearing in the training sets. However, the few-shot scenario needs models to classify objects with few instances without any extra information. Recently, \citet{yu2018diverse} propose a multi-metric method for few-shot text classification. However, there lack systematic researches about adopting few-shot learning for NLP tasks. We propose {\dataset}: a new large-scale supervised \textbf{Few}-shot \textbf{Rel}ation Classification dataset. To address the wrong labeling problem in most distantly supervised RC datasets, we apply crowd-sourcing to manually remove the noise.\footnote{Many previous works, such as \citep{roth2013survey,luo2017learning,xin2018put} have worked on automatically removing noise from distantly supervision. Instead, we use crowd-sourcing methods to achieve a high accuracy.} 

Besides constructing the dataset, we systematically implement the most recent state-of-the-art few-shot learning methods and adapt them for RC. We conduct a detailed evaluation for all these models on our dataset. Though the state-of-the-art few-shot learning methods have much lower results than humans on our challenging dataset, they significantly outperform the vanilla RC models, indicating that incorporating few-shot learning is promising and needs further research. In summary, our contribution is three-fold: 

(1) We formulate RC as a few-shot learning task, and propose a new large supervised few-shot RC dataset. 

(2) We systematically adapt the most recent state-of-the-art few-shot learning methods for RC, which may further benefit other NLP tasks. 

(3) We conduct a comprehensive evaluation of few-shot learning methods on our dataset, which indicates some promising research directions for RC.

\section{{\dataset} Dataset}
In this section, we describe the process of creating {\dataset} in detail.
The whole procedure can be divided into two steps: (1) We create a large candidate set of sentences aligned to relations via distant supervision. (2) We ask human annotators to filter out the wrong labeled sentences for each relation to finally achieve a clean RC dataset.

\subsection{Distant Supervision}
For the first step, We use Wikipedia as the corpus\footnote{We use whole Wikipedia articles as corpus, not just the first sentence.} and Wikidata as the KB. Wikidata is a large-scale KB where many entities are already linked to Wikipedia articles. The articles in Wikipedia also contain anchors linking to each other. Thus it is convenient to align sentences in Wikipedia articles to KB facts in Wikidata. We also employ entity linking technique to extract more unanchored entities in articles. We first adopt named entity recognition via spaCy\footnote{\url{https://spacy.io/}} to find possible entity mentions, then match each mention with the name of an entity in KBs, and link the mention to the entity if successfully matched.

For each sentence $s$ in Wikipedia articles containing head and tail entities $e_1$ and $e_2$, if there exists a Wikidata statement $(e_1, e_2, r)$ meaning $e_1$ and $e_2$ have the relation $r$, we denote the $(s, e_1, e_2, r)$ tuple as an instance and add it to the candidate set. Empirically, many instances of a given relation contain the same entity pair. For such relation, classifiers may prefer memorizing the entity pairs in the training instances rather than grasping the sentence semantics. Therefore, in the candidate set of each relation, we only keep $1$ instance for each unique entity pair. Finally, we remove relations with fewer than $1000$ instances, and randomly keep $1000$ instances for the rest of the relations. As a result, we get a candidate set of $122$ relations and $122,000$ instances. 

\subsection{Human Annotation}
Next, we invite some well-educated annotators to filter the raw data on a platform similar to Amazon MTurk developed by ourselves. The platform presents each annotator with one instance each time, by showing the sentence, two entities in the sentence, and the corresponding relation labeled by distant supervision. The platform also provides the name of the entities and relation in Wikidata accompanied with the detailed description of that relation. Then the annotator is asked to judge whether the relation could be deduced only from the sentence semantics. We also ask the annotator to mark an instance as negative if the sentence is not complete, or the mention is falsely linked with the entity.

Relations are randomly assigned to annotators from the candidate set, and each annotator will consecutively annotate $20$ instances of the same relation before switching to next relation. To ensure the labeling quality, each instance is labeled by at least two annotators. If the two annotators have disagreements on this instance, it will be assigned to a third annotator. As a result, each instance has at least two same annotations, which will be the final decision. After the annotation, we remove relations with fewer than $700$ positive instances. For the remaining $105$ relations, we calculate the inter-annotator agreement for each relation using the free-marginal multirater kappa \cite{randolph2005free}, and keep the top $100$ relations.


\subsection{Dataset Statistics}

The final {\dataset} dataset consists of $100$ relations, each has $700$ instances. A full list of relations, including their names and descriptions, is provided in Appendix A.2. The average number of tokens in each sentence is $24.99$, and there are $124,577$ unique tokens in total. Following recent meta-learning tasks \cite{Vinyals2016MatchingNF-MiniImageNet}, which use separate sets of classes for training and testing, we use $64$, $16$, and $20$ relations for training, validation, and testing respectively. Table \ref{tab:three-sets} provides a comparison of our {\dataset} dataset to two other popular few-shot classification datasets, Omniglot and mini-ImageNet. Table \ref{tab:re-sets} provides a comparison of {\dataset} to the previous RC datasets, including SemEval-2010 Task 8 dataset \citep{Hendrickx2009SemEval2010T8}, ACE 2003-2004 dataset \citep{Strassel2008LinguisticRA}, TACRED dataset \citep{Zhang2017PositionawareAA}, and NYT-10 dataset \citep{Riedel2010ModelingRA}. While some RC datasets contain instances with no relations (negative), we ignore such instances for comparison.

\begin{table}[t]
\centering
\small
\begin{tabular}{l|r|r|r}
\toprule
 Dataset & \#cls. & \#inst./cls & \#insts. \\
\midrule
Omniglot & $1,623$ & $20$ & $32,460$ \\
mini-ImageNet & $100$ & $600$ & $60,000$ \\
{\dataset} & $100$ & $700$ & $70,000$ \\
\bottomrule
\end{tabular}
\caption{Comparison of {\dataset} with Omniglot and mini-ImageNet.}
\label{tab:three-sets}
\end{table}

\begin{table}[t]
\centering
\small
\begin{tabular}{l|r|r}
\toprule
 Dataset & \#cls. & \#insts. \\
\midrule
SemEval-2010 Task 8 & $9$ & $6,674$ \\
ACE 2003-2004 & $24$ & $16,771$ \\
TACRED & $42$ & $21,784$ \\
NYT-10 & $57$ & $143,391$ \\
{\dataset} & $100$ & $70,000$ \\
\bottomrule
\end{tabular}
\caption{Comparison of {\dataset} with existing RC datasets. Note that negative (no relation) instances in some datasets are ignored.}
\label{tab:re-sets}
\end{table}

\section{Experiments}
\begin{table*}[!ht]
\centering
 \scalebox{0.85}{
\begin{tabular}{l|c|c|c|c}
\toprule
 Model         &    5 Way 1 Shot    &   5 Way 5 Shot     &  10 Way 1 Shot     &   10 Way 5 Shot        \\ 
\midrule
 Finetune (CNN)  & $44.21\pm0.44$  & $68.66\pm0.41$ & $27.30\pm0.28$ & $55.04\pm0.31$ \\ 
 Finetune (PCNN) & $45.64\pm0.62$  & $57.86\pm0.61$ & $29.65\pm0.40$ & $37.43\pm0.42$ \\
 kNN (CNN)     & $54.67\pm0.44$  & $68.77\pm0.41$ & $41.24\pm0.31$ & $55.87\pm0.31$ \\
 kNN (PCNN)   & $60.28\pm0.43$  & $72.41\pm0.39$ & $46.15\pm0.31$ & $59.11\pm0.30$ \\ 
\midrule
 Meta Network (CNN)  & $64.46\pm0.54$  & $80.57\pm0.48$ & $53.96\pm0.56$  & $69.23\pm0.52$ \\
 GNN (CNN)          & $66.23\pm0.75$  & $81.28\pm0.62$ & $46.27\pm0.80$ & $64.02 \pm0.77$  \\
 SNAIL (CNN)     & $67.29\pm 0.26$  & $79.40\pm0.22$ & $53.28\pm0.27$ &  $68.33\pm0.25$ \\
 Prototypical Network (CNN)& $69.20\pm 0.20$ & $84.79\pm0.16$ &  $56.44\pm0.22$& $75.55\pm0.19$ \\
\midrule
Human performance & $92.22\pm5.53$ & - & $85.88\pm7.40$ & - \\
\bottomrule
\end{tabular}
 }
\caption{Accuracies (\%) of all models on {\dataset} under four different settings.}
\label{tab:eval}
\end{table*}

We conduct comprehensive evaluations of vanilla RC models with simple strategies such as finetune or kNN on our new dataset. We also evaluate the recent state-of-the-art few-shot learning methods.

\subsection{Task Formulation}
In few-shot relation classification, we intend to obtain a function $F:(\mathcal{R}, \mathcal{S}, x)\mapsto y$. Here $\mathcal{R}=\{r_1, \ldots,r_m\}$ defines the relations that the instances are classified into. $\mathcal{S}$ is a support set 
\begin{equation}
\begin{aligned}
\mathcal{S}=\{&(x_1^1, r_1), (x_1^2, r_1), \ldots,(x_1^{n_1}, r_1),\\
&\ldots,\\
&(x_m^1, r_m), (x_m^2, r_m), \ldots,(x_m^{n_m}, r_m)
\}
\end{aligned}
\end{equation}
including $n_i$ instances for each relation $r_i\in\mathcal{R}$. For relation classification, a data instance $x_i^j$ is a sentence accompanied with a pair of entities. The query data $x$ is an unlabeled instance to classify, and $y\in\mathcal{R}$ is the prediction of $x$ given by $F$.

In recent research on few-shot learning, $N$ way $K$ shot setting is widely adopted. We follow this setting for the few-shot relation classification problem. To be exact, for $N$ way $K$ shot learning 
\begin{equation}
N=m=|\mathcal{R}|, K=n_1=\ldots=n_m
\end{equation} 

\subsection{Experiment Settings}

We consider four types of few-shot tasks in our experiments: 5 way 1 shot, 5 way 5 shot, 10 way 1 shot, 10 way 5 shot. Under this setting, we evaluate different few-shot training strategies and state-of-the-art few-shot learning methods built upon two widely used instance encoders, CNN \cite{Zeng2014RelationCV} and PCNN \cite{zeng2015distant}.

For both CNN and PCNN, the sentence is first represented to the input vectors by transforming each word into concatenation of word embeddings and position embeddings. In CNN, the input vectors pass a convolution layer, a max-pooling layer, and a non-linear activation layer to get the final output sentence embedding. PCNN is a variant of CNN, which replaces the max-pooling operation with a piecewise max-pooling operation.


To evaluate this two vanilla models in few-shot RC task, we first consider two training strategies, namely Finetune and kNN. For the Finetune baseline, it learns to classify all relations on the training set with CNN/PCNN, and tune parameters on the support set. We only tune the parameters of output layer, and keep other parameters unchanged. For the kNN baseline, it also jointly classifies all relations during training, while at the test time, it uses the neural networks to embed all the instances and then adopts k-nearest-neighbor (kNN) to classify the test instances.

By adapting them to relation classification, we also evaluate four recently proposed few-shot learning methods, including Meta Network \cite{Munkhdalai2017MetaN}, GNN \cite{garcia2018fewshotGNN}, SNAIL \cite{mishra2018SNAIL}, and Prototypical Network \cite{Snell2017PrototypicalNF}. We describe briefly about these baselines in Sec. \ref{sec:baselines}. If you are familiar with these methods, you can safely skip that subsection. The hyperparameters of each model are selected via grid search against the validation set.

Human performance is also evaluated under 5 way 1 shot setting and 10 way 1 shot setting. A human labeler is given $5/10$ instances from different relations and one extra test instance. Human labelers are asked to decide which relation the test instance belongs to. Note that these labelers are not provided the name of the relations and any extra information. Since 5 way 5 shot and 10 way 5 shot settings are easier, we only evaluate performance of 5 way 1 shot and 10 way 1 shot.

\subsection{Baselines of Few-shot Learning Models}
\label{sec:baselines}
\paragraph{Meta Network}
Meta Network \cite{Munkhdalai2017MetaN} is a meta learning algorithm utilizing a high level \textit{meta learner} on top of the traditional classification model, or \textit{base learner}, to supervise the training process. The weights of base learner are divided into two groups, fast weights and slow weights. Fast weights are generated by the meta learner, whereas slow weights are simply updated by minimizing classification loss. The fast weights are expected to help the model generalize to new tasks with very few training instances. 

\paragraph{GNN}
GNN \cite{garcia2018fewshotGNN} tackles the few-shot learning problem by considering each supporting instance or query instance as a node in the graph. For those instances in the support sets, label information is also embedded into the corresponding node representations. Graph neural networks are then employed to propagate the information between nodes. A query instance is expected to receive information from support sets in order to make the classification. In our adaption, while the instances are encoded by CNNs, labels are represented by one-hot encoding. 

\paragraph{SNAIL}
SNAIL \cite{mishra2018SNAIL} is a meta learning model that utilizes temporal convolutional neural networks and attention modules for fast learning from past experience. SNAIL arranges all the supporting instance-label pairs into a sequence and appends the query instance behind them. Such an order agrees with the temporal order of learning process where we learn information by reading supporting instances before making predictions for unlabeled instances. Temporal convolution (a 1-D convolution) is then performed along the sequence to aggregate information across different time steps and a causally masked attention model is used over the sequence to aggregate useful information from former instances to latter ones.

\paragraph{Prototypical Networks}
Prototypical Network \cite{Snell2017PrototypicalNF}  is a few-shot classification model based on the assumption that for each class there exists a prototype. The model tries to find the prototypes for classes from supporting instances, and compares the distance between the query instance and each prototype under certain distance metric. Prototypical network learns a embedding function $u$ to embed each class's instances, and computes each prototype by averaging over all the output embeddings of instances in the support set $\mathcal{S}$ that are labeled with the corresponding class.

\section{Result Analysis and Future Work}

We report evaluation results in Table \ref{tab:eval}. From our preliminary experiments, PCNN with few-shot learning methods perform 3-10 percentages worse than CNN, therefore only CNN results are shown in our experimental results. From the results, we observe that integrating few-shot learning methods into CNN significantly outperforms CNN/PCNN with finetune or kNN, which means adapting few-shot learning methods for RC is promising. However, there are still huge gaps between their performance and humans', which means our dataset is a challenging testbed for both relation classification and few-shot learning.

\begin{table}[t]
\centering
\small
\scalebox{0.9}{
\begin{tabular}{m{0.75\columnwidth}|m{0.2\columnwidth}}
\toprule
Sentence & Reasoning \\ \midrule
\textcolor{blue}{\emph{Chris Bohjalian}} graduated from \textcolor{red}{\emph{Amherst College}} Summa Cum Laude, where he was a member of the Phi Beta Kappa Society. & Simple Pattern\\ \hline
\textcolor{blue}{\emph{James Alty}} obtained a 1st class honours (Physics) at \textcolor{red}{\emph{Liverpool University}}. & Common-sense Reasoning \\ \hline
He was a professor at \textcolor{red}{\emph{Reed College}}, where he taught \textcolor{blue}{\emph{Steve Jobs}}, and replaced Lloyd J. Reynolds as the head of the calligraphy program. & Logical Reasoning \\ \hline
He and \textcolor{blue}{\emph{Cesare Borgia}} were thought to be close friends since childhood, going on to accompany one another during their studies at the \textcolor{red}{\emph{University of Pisa.}} & Co-reference Reasoning \\
\bottomrule

\end{tabular}}
\caption{Examples from relation ``educated\_at''. Different colors indicate different entities, \textcolor{blue}{blue} for head entity, and \textcolor{red}{red} for tail entity.}
\label{tab:examples}
\end{table}
In this paper, we propose a new large and high quality dataset, {\dataset}, for few-shot relation classification task. This dataset provides a new point of view for RC, and also a new benchmark for few-shot learning. Through the evaluation of different few-shot learning methods, we find even the best model performs much worse than humans, which suggests there is still large space for few-shot learning methods to improve.

The most challenging characteristic of our dataset is the diversity in expressing the same relation. We provide some examples from {\dataset} in Table \ref{tab:examples}, showing different reasoning modes needed for classifying some instances. Future researches may consider incorporating commonsense knowledge or improved causal modules.

\section*{Acknowledgement}
This work is supported by the National Natural Science Foundation of China (NSFC No. 61572273, 61532010). This work is also funded by the Natural Science Foundation of China (NSFC) and the German Research Foundation (DFG) in Project Crossmodal Learning, NSFC 61621136008 / DFC TRR-169. Hao Zhu is supported by Tsinghua University Initiative Scientific Research Program. We thank all annotators for their hard work. We also thank all members from Tsinghua NLP Lab for their strong support for annotator recruitment. 

\bibliography{emnlp2018}
\bibliographystyle{acl_natbib_nourl}

 \end{document}